\documentclass[conference]{IEEEtran}
\IEEEoverridecommandlockouts
\usepackage[utf8]{inputenc}
\usepackage{cite}

\usepackage{amsmath,amssymb}
\usepackage{xfrac}
\usepackage{dsfont}
\usepackage{amsthm}
\newtheorem*{remark}{Remark}

\usepackage{booktabs}
\usepackage{multirow}
\usepackage[left=2.5cm, right=2.5cm]{geometry}
\usepackage[width=.\linewidth, font=small]{caption}
\usepackage{tikz,pgfplots,pgf}
\usepackage{tikzscale}
\pgfplotsset{compat=1.14}
\usepgfplotslibrary{fillbetween}
\usetikzlibrary{arrows.meta, fit}
\usetikzlibrary{shapes.multipart}
\usetikzlibrary{shapes, shapes.geometric, calc, arrows, snakes}
\usetikzlibrary{patterns,backgrounds}
\usepackage{color}
\usepackage{listings}
\usepackage{hyperref}
\usepackage{todonotes}
\usepackage{ifthen}
\DeclareMathOperator*{\argmax}{arg\,max}
\definecolor{color1}{HTML}{AECEA1}
\definecolor{color2}{HTML}{488389}
\definecolor{color3}{HTML}{383C65}
\definecolor{color4}{HTML}{2B1F3E}
\let\pmm\pm
\renewcommand{\pm}{\!\pmm\!}

\title{Noisy Labels for Weakly Supervised Gamma~Hadron~Classification
}

\author{
  \IEEEauthorblockN{Lukas Pfahler, Mirko Bunse, Katharina Morik}
  \IEEEauthorblockA{
  \textit{Artificial Intelligence Group} \\
  \textit{TU Dortmund University}\\
  Dortmund, Germany \\
  firstname.lastname@tu-dortmund.de}
}

\date{\today}
\begin{document}

\maketitle

\begin{abstract}

Gamma hadron classification, a central machine learning task in gamma ray astronomy, is conventionally tackled with supervised learning. However, the supervised approach requires annotated training data to be produced in sophisticated and costly simulations.
We propose to instead solve gamma hadron classification with a noisy label approach that only uses unlabeled data recorded by the real telescope.
To this end, we employ the significance of detection as a learning criterion which addresses this form of weak supervision.
We show that models which are based on the significance of detection deliver state-of-the-art results, despite being exclusively trained with noisy labels; put differently, our models do not require the costly simulated ground-truth labels that astronomers otherwise employ for classifier training.
Our weakly supervised models exhibit competitive performances also on imbalanced data sets that stem from a variety of other application domains.
In contrast to existing work on class-conditional label noise, we assume that only one of the class-wise noise rates is known.
%
\end{abstract}

\section{Introduction}
Supervision in machine learning is traditionally provided in the form of ground-truth labels that are assigned to every instance of a training data set. From this true assignment, a model can learn to predict a label for any other instance it is later presented with. But what if true labels are missing? Research on machine learning has explored a multitude of settings \cite{hernandez2016weak} with weaker forms of supervision, e.g. noisy labels \cite{menon2018learning,natarajan2013learning,scott2013classification,ghosh2015making}, (partial) equivalence relations \cite{wagstaff2001constrained,kumar2007classification}, zero-shot \cite{Romera-Paredes/Torr/2015} or few-shot learning \cite{wang2020generalizing}, multiple-instance learning \cite{foulds2010review,dietterich1997solving}, and learning from label proportions \cite{Stolpe/Morik/2011a}, just to name a few.

Modern gamma ray astronomy addresses its lack of labels differently: the field relies on sophisticated simulations as alternative ground-truth data providers \cite{Bockermann/etal/2015a,anderhub2013design}. However, these simulations are computationally demanding; reducing the amount of simulated training data yields considerable savings of time and energy.

In this paper, we explore the potential of weak supervision for \emph{gamma hadron classification} \cite{bock2004methods,Buschjaeger/etal/2020a}, a central learning task in gamma ray astronomy. By learning directly from the unlabeled data that a telescope produces anyway, we solve this task without the need for costly simulations. Namely, we learn a strong classifier by adopting one of the features, namely the origin of gamma and hadron particles, as a noisy weak label.

Discussing our approach within the theoretical framework of noisy labels even reveals a more general contribution of our work: unlike existing approaches in noisy label learning \cite{natarajan2013learning,scott2013classification,ghosh2015making}, we do not require any knowledge about the class-wise noise rates.
In fact, all we assume is that the noisy-positive class contains more truly-positive instances than the noisy-negative class. Based on this weak assumption, we propose an interpretable optimization criterion for classifier training in general. 

Sec.~\ref{sec:background} presents our contributions against the background of noisy label learning. Gamma ray astronomy is detailed in  Sec.~\ref{sec:physics}. Sec.~\ref{sec:algorithms} proposes two algorithms for noisy supervision: we either cluster all unlabeled instances and make a post-hoc assignment from clusters to classes or we split the feature space via decision tree induction. Sec.~\ref{sec:experiments} shows that these weakly supervised algorithms compete with the strongly supervised state of the art. We connect our findings to other weak learning paradigms in Sec.~\ref{sec:related} and conclude with Sec.~\ref{sec:conclusion}.

\section{Background: Learning Under Class-Conditional Label Noise}
\label{sec:background}
In learning under random label noise, we only have access to training labels $\widehat{y} \in \{+1, -1\}$ that are noisy in the sense of being randomly flipped versions of the ``clean'' ground-truth labels $y \in \{+1, -1\}$. In particular, the class-conditional random label noise model (CCN) \cite{menon2015learning,natarajan2013learning,scott2013classification,blum1998combining} states that the labels are flipped according to probabilities $p_+$ and $p_-$ that depend on the true class $y$, but not on the features. Feasible learning under CCN is conventionally established by assuming that the noisy labels are correct on average, i.e. that
\begin{equation}
  p_+ + p_- < 1,
    \quad\text{where}\quad
    p_i = \mathbb P(\widehat{Y} = -i \mid Y = i).
    \label{eq:ccn}
\end{equation}

Theoretic studies \cite{menon2015learning,natarajan2013learning,scott2013classification,blum1998combining} have shown that even an \emph{optimal} classifier can be learned under CCN. Remarkably, the only difficulty lies in finding the decision threshold $\theta \in \mathbb R$; an optimal scoring function $h : \mathcal X \rightarrow \mathbb R$, which is to be thresholded at $\theta$, can be learned directly from the noisy labels.

\begin{remark}
  Assume that we knew the true probability density $\mathbb P(X \mid \widehat{Y})$ of the features conditioned on the noisy labels. According to Scott et al. \cite[see Proposition~1]{scott2013classification}, there is a one-to-one correspondence between noisy thresholds $\theta\in\mathbb R$ and clean thresholds $\lambda\in\mathbb R$, i.e. \mbox{$\forall\,x\in\mathcal X$} \,\mbox{$\forall\,\lambda\geq 0$} \,\mbox{$\exists\;\theta\geq 0$} such~that

  \begin{displaymath}
    \frac{\mathbb P(x \mid Y=+1)}{\mathbb P(x \mid Y=-1)} > \lambda
      \;\Leftrightarrow\;
      \frac{\mathbb P(x \mid \widehat{Y}=+1)}{\mathbb P(x \mid \widehat{Y}=-1)} > \theta.
  \end{displaymath}

  Due to this one-to-one correspondence, noisy and clean classifiers have same receiver operator characteristic (ROC) if they are based on the true class-conditional densities of the clean and noisy data. Depending on the data and the model, we might even be able to learn a good approximation of $\mathbb P(X \mid \widehat{Y})$ solely from noisy data; all we need to do then is to find a decision threshold that is optimal with respect to the clean labels.
\end{remark}

In fact, the area under the ROC curve is immune to CCN \cite{menon2015learning}, also when the classification model is not based on estimates of the densities $\mathbb P(X \mid \widehat{Y})$.

\subsection{Finding the Optimal Decision Threshold under CCN}
Other measures, such as accuracy, can be optimized in two steps: first, we fit a scoring function from the noisy labels and then set the decision threshold to a value that is optimal in terms of the clean labels \cite{menon2015learning}. Analogously, we can employ a label-dependent weighting of the loss function already during training \cite{natarajan2013learning,ghosh2015making}. However, finding an optimal threshold or optimal loss weights requires precise knowledge of the noise rates $p_+$ and $p_-$.

If $p_+$ and $p_-$ are unknown, we can resort to estimating them from the noisy data \cite{menon2015learning,scott2013classification}. Without access to clean labels, this approach requires additional assumptions, such as the existence of clean labels at a point in feature space where the other clean class has zero probability \cite{menon2015learning}. In the context of gamma ray astronomy, we conceive such assumptions as being overly restrictive. Under class imbalance, the estimation of noise rates becomes even more difficult \cite{mithal2017rapt}.

Alternatively, if we had access to a small amount of clean labels, we could tune the decision threshold directly on this clean set \cite{blum1998combining}. However, cleanly labeled real data is not available to gamma ray astronomers.

\subsection{Our Contributions}
For gamma hadron classification, astronomers have proposed to determine the decision threshold $\theta$ of a scoring function by maximizing a \emph{criterion} that resembles a hypothesis test.
In the more general scope of CCN, we demonstrate several qualities of this strategy:
\begin{itemize}
  \item straightforward: choosing $\theta$ by computing and maximizing the criterion is easy with standard optimization tools; noise rates do not need to be estimated.
  \item interpretable: the criterion resembles a hypothesis test which tells us whether learning is actually feasible with the given noisy labels.
  \item general: we assume $p_-$ to be known, but do not introduce assumptions otherwise needed \cite{menon2015learning} to estimate an unknown $p_+$ from noisy data.
  \item versatile: the criterion is not limited to finding $\theta$; we propose learning algorithms which simultaneously fit a scoring function and a threshold (i.e. a complete classifier) through maximizing the criterion.
\end{itemize}

Adopting a term from astronomy, we call this criterion the \emph{significance of detection}. 

\section{Classification in Gamma Ray Astronomy}
\label{sec:physics}
Modern gamma-ray astronomy utilizes imaging air Cherenkov telescopes \cite{actis2011design,tridon2010magic,anderhub2013design} to gather information about cosmic gamma ray emitters, e.g. supernova remnants and active galactic nuclei.
These telescopes record the interactions of cosmic gamma rays within Earth's atmosphere. The properties of the gamma rays, and therefore some important properties of their cosmic origin, can be reconstructed from these recordings.

Machine learning plays a crucial role in the reconstruction of the telescope data \cite{Bockermann/etal/2015a,Noethe/2020a}, as it tackles the assignment of particle properties (labels) to telescope recordings (instances).
One fundamental prediction task is to separate the interesting gamma ray observations from \emph{hadronic particle} observations \cite{bock2004methods,Buschjaeger/etal/2020a}. This binary classification problem emerges from the fact that imaging air Cherenkov telescopes are not only triggered by gamma rays; hadronic particles exhibit a similar behavior in the atmosphere and are therefore recorded, too.
Consequently, they must be filtered out of the data sample, to not aggravate the actual analysis of cosmic gamma ray sources.

The state of the art in this \emph{gamma hadron classification} task is to learn a classifier with strong supervision, i.e. with large amounts of simulated data including ground-truth labels. Here, the need for simulations arises from the fact that the real telescope recordings do not have ground-truth labels and manually labeling the recordings is not feasible. To ensure valid prediction models, astronomers are setting high standards for the simulation: it must resemble the entire detection process, from cascading particle interactions within the atmosphere up to artifacts from the telescope hardware and camera electronics. Consequently, running these simulations is time- and energy-consuming. Moreover, closing the gap between simulated and real telescope data requires to manually fine-tune countless parameters of the simulator, an undertaking that is prone to errors.

\subsection{Detection of Gamma Ray Emitters}
One purpose of gamma hadron classifiers is to detect the cosmic sources of gamma radiation: if the telescope keeps pointing at a sky coordinate where a gamma ray source is assumed, does it ``see'' this gamma ray source? Seeing sources in the data is not only aggravated by the presence of hadronic particles, but also by the presence of a cosmic gamma ray background. In fact, background gamma rays can originate from every position in the sky. To detect a gamma ray emitter therefore means to confirm that this emitter produces a rate of gamma rays that outstands the background rate of gamma radiation.

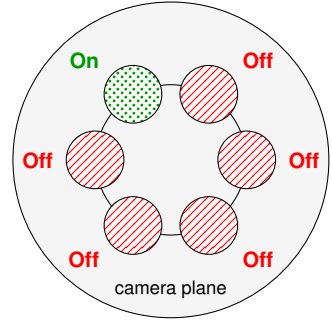
\begin{figure}
  \centering
  \begin{tikzpicture}[
  every node/.append style={font=\sffamily\footnotesize}
]
  
\def\drawconnectverticeslabelled[num vertex=#1, circle radius=#2, shift angle=#3] at (#4);{%
  \pgfmathtruncatemacro\vertices{#1}
  \pgfmathsetmacro\circleradius{#2}
  \draw[fill=black!4] (#4) circle (1.05*\circleradius cm);
  \node[anchor=south, font=\sffamily\scriptsize] at ([yshift=4pt]0,-1.05*\circleradius) {camera plane};
  \draw (#4) circle (1/2*\circleradius cm) node[regular polygon, regular polygon sides=\vertices, minimum size=\circleradius cm, draw=none, name={vertex set}] {};
  \foreach \x in {1,...,\vertices}{
    \ifthenelse{\equal {\x}{2}}{
      \draw[preaction={fill=white}, pattern=crosshatch dots, pattern color=green!60!black] (vertex set.corner \x) circle[radius=\circleradius*1.15/6] {};
      \pgfmathparse{#3-360*(\x)/ \vertices}
      \node at ($(vertex set)+(\pgfmathresult:\circleradius*2.8/5)$)[label={[font=\sffamily\footnotesize\bfseries, inner xsep=9pt]\pgfmathresult: \textcolor{green!60!black}{On}}]{};
    }{
      \draw[preaction={fill=white}, pattern=north east lines, pattern color=red] (vertex set.corner \x) circle[radius=\circleradius*1.15/6] {};
      \pgfmathparse{#3-360*(\x)/ \vertices}
      \node at ($(vertex set)+(\pgfmathresult:\circleradius*2.8/5)$)[label={[font=\sffamily\footnotesize\bfseries, inner xsep=9pt]\pgfmathresult: \textcolor{red}{Off}}]{};}
  }
}

\drawconnectverticeslabelled[num vertex=6, circle radius=2, shift angle=240] at (0,0);

\end{tikzpicture}
  \caption{%
    Illustration of the \emph{Wobble} mode for source detection.
    This mode observes the assumed source position, i.e. the \emph{On} region, simultaneously with multiple background regions. The FACT telescope \cite{anderhub2013design} employs five background regions, but other segmentations of the camera plane are also feasible.
  }
  \label{fig:wobble}
\end{figure}

An elegant way of simultaneously observing a gamma ray source and its background is to operate the telescope in the \emph{Wobble} mode \cite{fomin1994new} from Fig.~\ref{fig:wobble}.
This default mode of operation places the source on a ring around the camera center. The region around this position is called the \emph{On} region; it measures the rate of gamma rays that originate in the assumed source.
The additional \emph{Off} regions on the same ring simultaneously measure the background rate, provided that no other gamma ray source falls into any of them.
The FACT telescope \cite{anderhub2013design}, for instance, places six regions on a $0.6^{\circ}$ ring inside its $4.5^{\circ}$ field of view.
Every 20 minutes, another region is used as the \emph{On} region, which causes the telescope to ``wobble'' from region to region while observing a fixed position in the sky.
A gamma ray source is successfully detected when the \emph{On} region exhibits a rate of gamma rays that is considerably larger than the average rate exhibited by the \emph{Off} regions.


\subsection{Significance of Detection}
Formally, source detection is established as a hypothesis test. This test aims at rejecting the null hypothesis that the \emph{On} region does \emph{not} exceed the background rate. The statistical significance of such a rejection can then be interpreted as a measure of certainty: if this value is above some threshold, we can declare a successful detection. Moreover, values that exceed this threshold by large indicate that less data would have been needed for a detection; if the data set is fixed, the significance of detection therefore also measures the \emph{detection efficiency} of the telescope and the data analysis.

In this paper, we employ the test statistic by Li and Ma \cite{Li/etal/1983a}. Let $n^\gamma_\mathrm{on}$ and $n^\gamma_\mathrm{off}$ denote the number of gamma rays which originate in the \emph{On} and \emph{Off} regions, respectively. To compare the average rates in these regions, we also need a scaling factor $\alpha = \sfrac{A_\mathrm{on}}{A_\mathrm{off}}$, which is the ratio of the \emph{On} area and all \emph{Off} areas. For instance, the five equi-sized background regions of the FACT telescope, see Fig.~\ref{fig:wobble}, induce a factor $\alpha = \sfrac{1}{5}$. The significance of detection is given by
\begin{align}
  f(n^\gamma_\mathrm{on}, n^\gamma_\mathrm{off}) = \bigg[&
         2n^\gamma_\mathrm{on} \cdot \ln\left( \frac{1+\alpha}{\alpha} \cdot \frac{n^\gamma_\mathrm{on}}{n^\gamma_\mathrm{on}+n^\gamma_\mathrm{off}} \right)
         \label{eq:lima} \\%
    +\,& 2n^\gamma_\mathrm{off} \cdot \ln\left( (1  + \alpha)          \cdot \frac {n^\gamma_\mathrm{off}}{n^\gamma_\mathrm{on}+n^\gamma_\mathrm{off}} \right)
  \bigg]^{1/2} \nonumber
\end{align}

This statistic measures the significance of detection in the unit of ``sigmas'' ($\sigma$). Values of five and more sigma, i.e. $f(n^\gamma_\mathrm{on}, n^\gamma_\mathrm{off}) > 5$, are usually considered successful detections in gamma ray astronomy.

Note that hadronic instances are completely ignored in Eq.~\ref{eq:lima}. This fact results in a trade-off between precision and recall of the gamma hadron classifier: if all instances were classified as hadrons, both $n^\gamma_\mathrm{on}$ and $n^\gamma_\mathrm{off}$ would be zero and so would be the statistic; if all instances were classified as gammas, it would be unlikely that the \emph{On} region contains significantly more positive events than the \emph{Off} regions, so that the resulting statistic would be low, too. In contrast, a perfect classifier would achieve high significance values, provided that the \emph{On} region contains a gamma ray source indeed.

This line of thought has motivated astronomers to leverage Eq.~\ref{eq:lima} in choosing the decision threshold of their gamma hadron classifiers. First, such a classifier is trained, usually with strong supervision, i.e. with a labeled training set that is provided by a costly simulation. Second, the decision threshold of the trained classifier $h : \mathcal X \rightarrow \mathbb R$ is chosen such that the value of Eq.~\ref{eq:lima} is maximized on real-world data, for which we know the regions of origin, but not the labels.

Formally, let $X = (\vec{x}_1, \dots, \vec{x}_n, \; \vec{x}_{n+1}, \dots, \vec{x}_{n+m})$ be the concatenation of $n$ unlabeled real-world instances from the \emph{On} region and $m$ instances from the \emph{Off} region. Choose the decision threshold $\theta^\ast$ such that
\begin{equation}
  \theta^\ast = \argmax_{\theta \in \mathbb R} \, f\left(
    \sum_{i=1}^n \mathds{1}_{h(\vec{x}_i) > \theta},
    \sum_{i=1}^m \mathds{1}_{h(\vec{x}_{n+i}) > \theta}
  \right)
  \label{eq:threshold-tuning}
\end{equation}

\subsection{Noisy Labels from On and Off Information}
We establish noisy label learning for gamma hadron classification by identifying the \emph{On} and \emph{Off} regions as noisy labels for gamma rays and hadronic particles. This identification is justified if the \emph{On} region pictures a gamma ray source: only then, we can expect a correspondence between clean labels $y$ and noisy labels $\widehat{y}$. Namely, we expect that the \emph{On} region (our noisy-positive class) contains more gamma rays (truly-positive instances) than the average \emph{Off} region (our noisy-negative class). To ensure that this assumption holds, we train with observations of known gamma ray sources. 
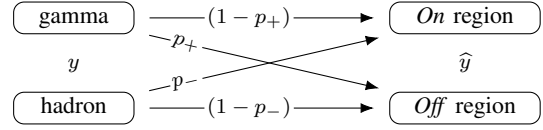
\begin{figure}
  \centering
  \begin{tikzpicture}[x=52mm, y=-12mm]

  \tikzstyle{box}=[
    draw=black,
    rounded corners,
    inner ysep=1pt,
    font=\small,
    text=black
  ]

  \tikzstyle{left box}=[box, minimum width=16mm]
  \tikzstyle{right box}=[box, minimum width=20mm]

  \tikzstyle{arrow}=[
    draw=black,
    -{Latex[length=1.8mm]},
    shorten <=2mm,
    shorten >=2mm
  ]

  \tikzstyle{arrow node}=[
    font=\footnotesize,
    text=black,
    sloped,
    fill=white,
    rounded corners,
    inner sep=1pt
  ]

  \node[left box] (gamma) at (0,0) {gamma\strut};
  \node[left box] (hadron) at (0,1) {hadron\strut};
  \node[right box] (on) at (1,0) {\emph{On} region\strut};
  \node[right box] (off) at (1,1) {\emph{Off} region\strut};

  \draw[arrow] (gamma) -- (on) node[arrow node, pos=.45] {$(1-p_+)$};
  \draw[arrow] (gamma) -- (off) node[arrow node, pos=.2] {$p_+$};
  \draw[arrow] (hadron) -- (on) node[arrow node, pos=.2] {$p_-$};
  \draw[arrow] (hadron) -- (off) node[arrow node, pos=.45] {$(1-p_-)$};

  \node[font=\footnotesize] at (0,.5) {$y$\strut};
  \node[font=\footnotesize] at (1,.5) {$\widehat{y}$\strut};

\end{tikzpicture}
  \caption{%
    The class-conditional random noise model $y \mapsto \widehat{y}$ in the context of weakly supervised gamma hadron classification. We employ the \emph{On} and \emph{Off} regions as noisy labels for gamma rays and hadronic particles with noise rates $p_+$ and $p_-$.
  }
  \label{fig:noise-model}
\end{figure}

Fig.~\ref{fig:noise-model} presents our noisy label interpretation of \emph{On} and \emph{Off} annotations with the resulting class-wise noise rates $p_+$ and $p_-$. In gamma ray astronomy, we are facing $p_- \approx \frac{A_\mathrm{on}}{A_\mathrm{on} + A_\mathrm{off}}$ in particular because the hadronic particles can originate in any sky position. Assuming that the \emph{On} region contains more gamma rays than the average \emph{Off} region translates to $p_+ < \frac{A_\mathrm{off}}{A_\mathrm{on} + A_\mathrm{off}}$, which nicely fits the standard CCN assumption from Eq.~\ref{eq:ccn}.
We employ our noisy labels in three distinct ways:
\begin{itemize}
  \item We follow existing work on CCN by learning a scoring function from the noisy labels, which is then thresholded via an estimation of the noise rates. In particular, we consider the thresholding techniques by Menon \cite{menon2015learning} and Mithal \cite{mithal2017rapt} in our experiments.
  \item We learn a scoring function from the noisy labels again, but set the threshold by maximizing the significance of detection. This approach is similar to the one from Eq.~\ref{eq:threshold-tuning}, with the difference that $h$ is trained with noisy labels rather than ground-truth.
  \item We learn a full classification model directly from the significance of detection, without separating the scoring function from the threshold. We develop the algorithms for this approach in the next section.
\end{itemize}


\section{Weakly-Supervised Algorithms based~on~the~Significance of Detection}
\label{sec:algorithms}
We propose two weakly supervised machine learning algorithms for binary classification under class-conditional label noise. Both algorithms learn by maximizing the significance of detection directly, thus overcoming the separation of the scoring function and the decision threshold that is usual in noisy label learning. While our algorithms are inspired by gamma hadron separation, they apply to any learning task that is aggravated by class-conditional label noise.

For generality, we omit the notion of the areas $A_\mathrm{on}$ and $A_\mathrm{off}$, which have defined $p_-$ and the $\alpha$ factor in the astronomy domain. In equivalence to assuming the values of such areas, we instead assume $p_-$ to be known. About $p_+$, contrastingly, we assume no knowledge. Re-arranging the area-based definitions of $\alpha$ and $p_-$ yields the $\alpha$ factor from Eq.~\ref{eq:lima} as
\begin{equation}
  \alpha \;=\; \frac{p_-}{1-p_-}.
\end{equation}

Remarkably, the hypothesis test from Eq.~\ref{eq:lima} is now able to detect violations of Eq.~\ref{eq:ccn}, the standard CCN assumption. Namely, if a significance outcome is small, say below a threshold of $5\sigma$ for example, we know that $p_+$ does not obey Eq.~\ref{eq:ccn} with sufficient certainty. In this case, learning from the given noisy labels is not feasible. This matter of interpretation is in contrast to existing methods \cite{menon2015learning,mithal2017rapt}, which take Eq.~\ref{eq:ccn} for granted.

\subsection{K-Means Detection}
Our first approach is centered around the k-means clustering algorithm, a standard unsupervised method in the toolbox of data science \cite{Hastie/etal/2003a}. The underlying assumption, as for supervised classification, is that the feature distributions of the two classes differ. Since clustering algorithms group similar feature vectors in the same clusters, some of these clusters are likely to contain an excess of one class or the other.

Let $X=\bigcup_{c=1}^k X_c$ with $X_c \cap X_{c'} = \emptyset$ for all $c \not = c'$ be the partitioning of all observations, as computed by $k$-means or any other cluster partitioning algorithm. Moreover, let $n_\mathrm{on}^{(c)}$ and $n_\mathrm{off}^{(c)}$ be the number of noisy-positive and noisy-negative instances in each cluster, respectively. We achieved satisfactory results with a standard $k$-means++ implementation from scikit-learn \cite{Pedregosa/etal/2011}, which applies a greedy initialization strategy \cite{Arthur/Vassilvitskii/2007a} and minimizes the squared Euclidean distance between the observations and the nearest of $k$ cluster centroids using Lloyd's algorithm. The only hyper-parameter of this approach is $k$, the number of clusters.

From such a partitioning, we compute the optimal assignment of clusters to \emph{clean} classes by maximizing the significance of detection from Eq.~\ref{eq:lima}. Namely, we find the optimal assignment $a^\ast \in \{0,1\}^k$, such that
\begin{equation}
  a^\ast =
	\argmax\limits_{a \in \{0,1\}^k} f\left(\sum\limits_{c=1}^k a_c n_\mathrm{on}^{(c)}, \, \sum\limits_{c=1}^k a_c n_\mathrm{off}^{(c)}\right).
  \label{eq:assignment}
\end{equation}

For computational efficiency, we relax this problem to a continuous, convex parameter space $a\in[0,1]^k$. This relaxation does not change the optimal solution $a^*$, but allows us to apply first-order constrained optimization methods to efficiently find the optimal assignment.


At inference time, we determine the class of an event $\vec{x}$ by first assigning it to its nearest cluster centroid, thereby determining its cluster $c(\vec{x})$. Then, we return the clean class $a^*_{c(\vec{x})} \in \{0,1\}$ that is associated with this cluster.

\subsection{Decision Tree Detection}
Our second approach builds on decision tree induction. We modify this popular method for supervised classification such that it directly maximizes the significance of detection with noisy-labeled training data.

Like classic decision trees, our algorithm works by recursively partitioning the training set in a greedy fashion. At each node, we split the data by comparing a single feature to a learned threshold. To find the best split, we test every possible value in the data and evaluate the significance of detection for both sides of the partition. When the value does not increase, as compared to not splitting, we return. Alternatively, we return when a user-defined maximum tree depth is reached.

We denote by $X[x_{\cdot j} \leq \theta]$ all examples in the current node where the $j$-th feature is less or equal $\theta$ and analogously define $X[x_{\cdot j} > \theta]$ the complement. For each candidate value of $j$ and $\theta$, we evaluate Eq.~\ref{eq:lima} to derive the optimal split as
\begin{align}
	j^*, \, \theta^* &= \argmax_{j, \theta} \; \max\left\{f_j^{\leq\theta}, \; f_j^{\vphantom\leq>\theta}\right\}, \label{eq:split}\\
	\nonumber\text{where} \enspace f_j^{\leq\theta} &= f\left(n_\mathrm{on}(X[x_{\cdot j} \leq \theta]), \, n_\mathrm{off}(X[x_{\cdot j} \leq \theta])\right) \\
	\nonumber\text{and} \enspace f_j^{>\theta} &= f\left(n_\mathrm{on}(X[x_{\cdot j} > \theta]), \, n_\mathrm{off}(X[x_{\cdot j} > \theta])\right).
\end{align}

The best split according to Eq.~\ref{eq:split} can be evaluated efficiently by sorting the data according to each of the features. Considering only the maximum of $f_j^{\leq\theta}$ and $f_j^{\vphantom\leq>\theta}$ amounts to the fact that a greedy maximization of Eq.~\ref{eq:lima} does not require balanced splits: if the other side of a split results in a low sigma value, we can either discard this side without harming the overall significance of detection or we can split this side further in a later step.

As an alternative to Eq.~\ref{eq:split}, we can conceive a slightly modified criterion that balances both sides of each split. It can be shown that the following alternative is equivalent to maximizing the traditional information-gain split criterion of decision trees with noisy labels:
\begin{equation}
	j^*, \, \theta^* = \argmax_{j, \theta} \; \left(f_j^{\leq\theta}\right)^2 + \left( f_j^{\vphantom\leq>\theta}\right)^2 \label{eq:split2}
\end{equation}


\subsection{Ensemble Variants}
Both algorithms lend themselves for application in an ensemble. We choose a bagging approach \cite{Breiman/1996a} as with classical random forests \cite{Breiman/2001a}. Instead of a single model, we compute $T$ models and employ randomization techniques to obtain $T$ different decision functions.  We use the following sources of randomization:
\begin{itemize}
  \item In the K-Means ensemble, we base each of the $T$ clusterings on a random subset of the features as well as a boostrapped sample of the events, i.e. we sample $n$ events with replacement.
  \item In the Decision-Forest ensemble, we use a bootstrapped sample of the events and choose the optimal splits only on a random subset of features of size $\lfloor\sqrt d \rfloor$ that we sample at each split.
\end{itemize}

We combine the $T$ models by returning the mean value of their classification outputs. 
To tune the decision threshold of the ensemble methods, we optimize Eq.~\ref{eq:threshold-tuning}.

\section{Experiments}
\label{sec:experiments}
In the following, we investigate the performance of our proposed methods in comparison with existing methods from CCN learning and in comparison with strongly supervised methods.
The implementation of our methods and experiments is published on GitHub\footnote{See \url{https://github.com/tudo-ls8/unsupervised_fact/}}.

\subsection{Validation Protocol for the Telescope Data}
\label{sec:validation}
Due to the lack of labeled real-world data in gamma ray astronomy, we can report the real-world performance of our models only in terms of their significance of detection, as according to Eq.~\ref{eq:lima}, but not in terms of supervised measures such as accuracy or F1 score. This matter also has an implication on the way we cross-validate our results: we repeatedly train and predict on disjoint splits of the data, as usual. However, we store the predictions of each holdout set, rather than computing separate scores for the folds. This way, we obtain predictions for all instances in the data, but each prediction is based on a model that has not seen the instance during training. The reported values of the significance of detection are then obtained with the full data set.

For reliable results, the training and hold-out sets of the data should be as independent as possible. Since recording conditions, like the brightness of the night sky, change during the course of data taking, we group the data based on an attribute that corresponds to the time of the day. We found that this grouping yields the lowest values of the significance of detection, giving us the most cautious estimates of model performances.

\subsection{Detection of the Crab Nebula}
We use the open data sample of the FACT collaboration.\footnote{\url{https://factdata.app.tu-dortmund.de/}}. This data contains five nights of wobble mode observations, totalling 757,993 instances, where the telescope is pointed at the Crab Nebula, a bright supernova remnant. Each instance in the dataset corresponds to a camera recording of an event. We use 22 handcrafted, high-level features extracted from the camera recordings\cite{Bockermann/etal/2015a}. These features include statistics like the number of pixels that recorded photons and geometric traits \cite{Hillas/1985} that describe the shape of the event on the camera surface. We expect that the vast majority of events are induced by hadronic particles and that only around 0.1\% to 1.0\% of them are gamma ray events. The data contains 4,332 \emph{On} events and 17,229 \emph{Off} events.

We compare our methods to two different approaches for learning with class-conditional label noise, namely to the approaches by Menon et al \cite{menon2015learning} and by Mithal et al \cite{mithal2017rapt}, the latter of which is designed for the particular case of imbalanced class frequencies. Both methods start by learning any standard classifier from weak labels, which is decision trees and random forests in our experiments. In a second step, both methods estimate the class-wise noise rates from the noisy-labeled data and adapt their decision thresholds according to these estimates.

We further establish a comparison to the gamma hadron predictions from the default and fully supervised pipeline\footnote{\url{https://github.com/fact-project/open_crab_sample_analysis}}. These predictions are produced by a random forest classifier that is trained on 240,000 simulated and ground-truth labeled instances with balanced classes. 
\begin{table}
\caption{Significance scores for the detection of the Crab Nebula (see Eq.~\ref{eq:lima}, higher is better). Our algorithms are compared to existing CCN methods and to the strongly supervised state-of-the-art. This table presents only the best performances; additional results are given in the supplementary material.}
\label{table:insample_results}
\centering
\addtolength{\tabcolsep}{-1pt}   
\begin{tabular}{@{}llcc@{}}
\toprule
\multicolumn{2}{l}{Model}         & Single Model & Ensemble $(T=100)$ \\ \midrule
\multicolumn{2}{l}{\textbf{k-Means} (Eq.~\ref{eq:assignment})}   & &          \\
         & k=48                   &\textbf{19.5799 $\pm$ 0.3708}&$23.1176 \pm 0.1974$\\
         & k=256                  &$17.7766 \pm 0.5713$&\textbf{24.3246 $\pm$ 0.4966}\\
\multicolumn{2}{l}{\textbf{LiMa Tree} (Eq.~\ref{eq:split})} & & \\
         & max\_depth=4           &\textbf{21.1666 $\pm$ 0.0000}&$23.3015 \pm 0.3321$\\
         & max\_depth=6           &$18.4227 \pm 0.0000$&\textbf{25.1085 $\pm$ 0.2236}\\
\multicolumn{2}{l}{\textbf{Noisy Tree} (Eq.~\ref{eq:split2})} & & \\         
         & max\_depth=4           &\textbf{24.2899 $\pm$ 0.0000}&$25.3015 \pm 0.3321$\\
         & max\_depth=8           &$21.0170 \pm 0.0000$&\textbf{26.3175 $\pm$ 0.2479}\\
\multicolumn{2}{l}{\textbf{Menon Tree} \cite{menon2015learning}} & & \\         
         & max\_depth=4           &\textbf{24.2899 $\pm$ 0.0000}&$25.2321 \pm 0.1812$\\
         & max\_depth=7           &$18.3020 \pm 0.0000$&\textbf{26.0165 $\pm$ 0.2950}\\
\multicolumn{2}{l}{\textbf{Mithal Tree} \cite{mithal2017rapt}} & & \\         
         & max\_depth=4           &\textbf{24.2899 $\pm$ 0.0000}&$25.0858 \pm 0.2089$\\
         & max\_depth=8           &$19.7144 \pm 0.0000$&\textbf{26.4465 $\pm$ 0.1804}\\
\multicolumn{2}{l}{\textbf{Supervised} (SOTA)} & & \\         
         & DT / RF                &\textbf{24.6506 $\pm$ 0.0000}&\textbf{26.2526$\pm$ 0.2024}\\
%
\bottomrule
\end{tabular}

\end{table}

We test a number of different hyper-parameters per model and repeat each setup ten times to report average values and standard deviations, as displayed in Table~\ref{table:insample_results}. We see that the noisy-label approach to gamma hadron classification with random forests performs as good as the traditional approach that is based on costly, simulated training data with full supervision. This remarkable result could shape the way in which future telescopes approach their analysis pipelines: it demonstrates that we can benefit from real recordings for training, not only as background events, as it has been done before \cite{Acciari/etal/2020}, but also for learning about the gamma class. 
Gamma hadron classification could become even more data driven than today and less sensitive to imperfections and biases that are exhibited by the simulations.

Once having decided to use noisy labels, the choice of the particular method for tuning the decision threshold is only a secondary concern, as all of the tested approaches yield high significance scores between 25.0 and 26.4 sigmas. The noisy random forest (Eq.~\ref{eq:split2}), Menon forest \cite{menon2015learning} and Mithal forest \cite{mithal2017rapt} even outperform the supervised model on average, albeit by a slim margin.

The plausibility of these scores is verified by another experiment in which we violate the assumption that the \emph{On} region contains more gamma events than any of the \emph{Off} regions. Namely, we artificially remove the \emph{On} region and declare one of the \emph{Off} regions as being the new \emph{On} region. All our methods now return approximately zero sigmas, which means that the violation of assumptions is successfully detected. The plausibility of the scores in Tab.~\ref{table:insample_results} is further supported by the observation that our weakly supervised predictions achieve high correlations with the strongly supervised state-of-the-art and high classification performances on simulated, ground-truth test data. We further detail these plausibility checks in the appendices~\hyperref[sec:apx:violated]{A} and~\hyperref[sec:apx:agreement]{B}.

%
%

\subsection{Results in Other Application Domains}
Our proposed methods are not limited to gamma ray astronomy. In fact, methods based on the significance of detection can be applied in any application domain where we expect substantially more true positives in the noisy-positive class than we expect in the noisy-negative class. We support this claim with experiments on standard imbalanced datasets where we artificially inject class-dependent noise into the labels. We focus on the setting $p_- = \frac{1}{2}$ and consequently set $\alpha=1$ in Eq.~\ref{eq:lima}. Hence, the null-hypothesis that we intend to reject is that the noisy-positive class and the noisy-negative class contain the same number of true positives.

In this experiment, we evaluate random forests that are trained on noisy labels and use 10-fold stratified cross-validation to estimate the F1 score. We optimize the decision thresholds on the out-of-bag predictions, such that we do not need a separate calibration dataset. Like before, we compare our significance-based thresholds to the thresholds that have been proposed earlier \cite{menon2015learning,mithal2017rapt}.
We try different levels of label noise, as presented in Tab.~\ref{table:classification_results} and in appendix~\ref{sec:apx:other}. We use all datasets available in the \texttt{imblearn} library \cite{lemaitre2017imbalanced} with at most 100 features and at least 200 samples for the minority class.

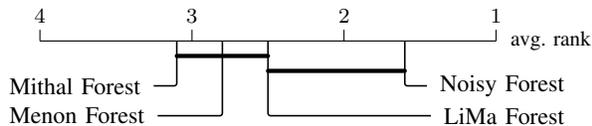
\begin{figure}[!b]
  \centering


\begin{tikzpicture}[]
  \begin{axis}[
    clip=false, axis x line=center, axis y line=none, xmin=1, xmax=4, ymin=-3.5, ymax=0, scale only axis, height=3.3\baselineskip, width=.75*\linewidth, ticklabel style={anchor=south, yshift=1.33*\pgfkeysvalueof{/pgfplots/major tick length}, font=\footnotesize}, every tick/.style={yshift=.5*\pgfkeysvalueof{/pgfplots/major tick length}}, axis line style={-}, title style={yshift=\baselineskip}, xtick={1,2,3,4}, xlabel={avg. rank}, xlabel style={anchor=west, xshift=2pt, font=\footnotesize}, x dir=reverse
  ]
  
  \draw[semithick, rounded corners=1pt] (axis cs:1.6, 0) |- (axis cs:0.55, -1.5) node[font=\small, fill=white, inner xsep=5pt, outer xsep=-5pt, anchor=east] {Noisy Forest};
  
  \draw[semithick, rounded corners=1pt] (axis cs:2.5, 0) |- (axis cs:0.55, -2.5) node[font=\small, fill=white, inner xsep=5pt, outer xsep=-5pt, anchor=east] {LiMa Forest};
  
  \draw[semithick, rounded corners=1pt] (axis cs:2.8, 0) |- (axis cs:4.2, -2.5) node[font=\small, fill=white, inner xsep=5pt, outer xsep=-5pt, anchor=west] {Menon Forest};
  
  \draw[semithick, rounded corners=1pt] (axis cs:3.1, 0) |- (axis cs:4.2, -1.5) node[font=\small, fill=white, inner xsep=5pt, outer xsep=-5pt, anchor=west] {Mithal Forest};
  
  \draw[ultra thick, line cap=round] (axis cs:2.5, -0.5) -- (axis cs:3.1, -0.5);
  
  \draw[ultra thick, line cap=round] (axis cs:1.6, -1.0) -- (axis cs:2.5, -1.0);
  
  \end{axis}
\end{tikzpicture}

  \caption{%
    Critical difference diagram \cite{demsar2006statistical}, based on Table~\ref{table:classification_results}, at a confidence level of 0.9. Methods are connected with horizontal bars if and only if a Holm-corrected Wilcoxon signed-rank test cannot significantly distinguish their pairwise performances.
  }
  \label{fig:ccd}
\end{figure}

We see that our two methods, which are based on significance-based thresholds, outperform the existing methods on 8 out of 10 datasets. Our forest based on Eq.~\ref{eq:split2} has the highest average F1 score. On the other datasets, our methods do not dramatically deviate from the winner, whereas both competitors exhibit some catastrophic failures. The critical difference diagram \cite{demsar2006statistical} in Fig~\ref{fig:ccd} confirms that the Noisy Forest significantly outperforms the existing two approaches.

\begin{table}
\caption{Averaged F1 scores (20 trials, higher is better) on imbalanced data sets with noise levels $p_+=0.1$ and $p_-=0.5$.}
\label{table:classification_results}
\centering
\addtolength{\tabcolsep}{-3pt}   
\scriptsize
\begin{tabular}{lcccc}
\toprule
Dataset        & LiMa Forest         & Noisy Forest        & Menon Forest         & Mithal Forest       \\ \midrule
satimage       & 0.563 $\pm$ 0.012   & \textbf{0.602 $\pm$ 0.010} & 0.549 $\pm$ 0.0055  & 0.563 $\pm$ 0.061 \\
proteinhomo    & 0.788 $\pm$ 0.002   & \textbf{0.790 $\pm$ 0.002} & 0.339 $\pm$ 0.0076  & 0.786 $\pm$ 0.002 \\
opticaldigits  & 0.813 $\pm$ 0.008   & \textbf{0.859 $\pm$ 0.008} & 0.843 $\pm$ 0.0051  & 0.826 $\pm$ 0.007 \\
pendigits      & \textbf{0.961 $\pm$ 0.006}   & 0.955 $\pm$ 0.002 & 0.945 $\pm$ 0.003  & 0.961 $\pm$ 0.003 \\
letterimg      & 0.793 $\pm$ 0.013   & 0.809 $\pm$ 0.007 & 0.809 $\pm$ 0.005  & \textbf{0.810 $\pm$ 0.007}\\
coil2000       & 0.115 $\pm$ 0.003   & 0.146 $\pm$ 0.004 & \textbf{0.160 $\pm$ 0.002}  & 0.114 $\pm$ 0.003 \\
thyroidsick    & 0.769 $\pm$ 0.015   & \textbf{0.786 $\pm$ 0.008} & 0.775 $\pm$ 0.008  & 0.400 $\pm$ 0.176 \\
sickeuthyroid  & \textbf{0.836 $\pm$ 0.006}   & 0.772 $\pm$ 0.011 & 0.701 $\pm$ 0.009  & 0.678 $\pm$ 0.136 \\
mamography     & 0.509 $\pm$ 0.023   & \textbf{0.569 $\pm$ 0.012} & 0.484 $\pm$ 0.011  & 0.103 $\pm$ 0.037 \\
abalone        & 0.370 $\pm$ 0.012   & \textbf{0.384 $\pm$ 0.006} & 0.359 $\pm$ 0.004  & 0.300 $\pm$ 0.038 \\[1pt]
avg. rank      & 2.500               & \textbf{1.600}      & 2.800                & 3.100              \\
avg. f1        &  0.658              & \textbf{0.667}      & 0.596                & 0.554              \\
\bottomrule
\end{tabular}
\end{table}

\section{Related Work}
\label{sec:related}
In this work, we have addressed class-conditional label noise (CCN) for weakly supervised binary classification. Since the background of this particular learning task is already detailed in Sec.~\ref{sec:background}, we focus on connections to other strands of weak supervision in the following.

Beyond the CCN noise model, other types of label noise have been discussed in the scientific literature. For instance, the label noise of a distribution is called \emph{uniform} \cite{ghosh2015making} if each label has the same chance of being flipped, independent of the class and of the instance. Otherwise, if the chance of being flipped does not depend on the true class, but on the features, we are speaking of \emph{purely instance-dependent} label noise \cite{menon2018learning}. Generally speaking, learning is feasible under each of these noise models. However, learning under each noise model requires dedicated assumptions about the data and the learning method, with CCN combining the most appropriate assumptions in the context of gamma hadron classification.
Since the distinction of gamma rays and hadronic particles naturally poses a \emph{binary} classification task, we have decided to leave the difficulties that are otherwise imposed by multi-class CCN \cite{patrini2017making} for future work on learning through the significance of detection.

The \emph{On} and \emph{Off} noisy labels considered here depict only a single source of weak supervision, i.e. a single noisy label is assigned to each training instance. If, however, multiple sources of weak labels were available, we could combine them \cite{ratner2020snorkel} to learn classifiers from less data \cite{zamani2018theory}. To this end, we see a potential for future work in gamma ray astronomy: domain experts might be able to define additional noisy labels, e.g. through heuristics that express physical background knowledge, that can be incorporated into the learning process.


\subsection{A label-free interpretation of noisy label learning}
Our \emph{On} and \emph{Off} annotations can alternatively be understood as a special group attribute rather than being understood as noisy labels. Indeed, these annotations are not only available for the training instances, but also for any test instance that is to be predicted.
In particle physics, this label-free interpretation of \emph{On} and \emph{Off} information is actually the prevalent one \cite{metodiev2017classification,cohen2018machine,desimone2019guiding}.
It allows us to draw additional connections of our work to multiple-instance (MI) learning \cite{foulds2010review,dietterich1997solving} and learning from label proportions (LLP) \cite{Stolpe/Morik/2011a}.

These strands of weak supervision challenge the traditional one-to-one correspondence between labels and instances. Namely, they assume that labels are only given for bags of instances but not for each instance individually. These bags can vary in size, which leads to a one-to-many correspondence between labels and instances. Similarly, equivalence relations \cite{wagstaff2001constrained,kumar2007classification} exhibit a one-to-two correspondence; they assume labels only for pairs of instances. In all of these settings, the goal is to learn a classifier that predicts the labels of individual instances, just like in standard supervised learning, despite the fact that no individual labels are available for training.

In the label-free interpretation of our approach,
all we know for supervision is that the significance of detection must be maximized. This maximization amounts to a one-to-all correspondence between supervision and instances which goes beyond MI learning and LLP: we are not learning from labels, but by maximizing a single domain-specific criterion over the \emph{full} training set.

Still, we have focused on the noisy label interpretation of \emph{On} and \emph{Off} annotations throughout this work. This focus is due to two reasons: to shed light on the theoretical properties of learning with \emph{On} and \emph{Off} information and to generalize our proposals to other CCN applications.

\subsection{Related applications from particle physics}
\emph{On} and \emph{Off} annotations are often available in data from particle detectors. Similar to our use case, the goal of these detectors is to distinguish some kind of signal (like our gamma rays) from some kind of background instances (like our hadronic particles). Traditionally, this distinction is learned with full supervision provided by simulations. A more recent proposal, however, is to learn directly from \emph{On} and \emph{Off} regions, instead  \cite{metodiev2017classification,cohen2018machine,desimone2019guiding}. 
Since this proposal essentially learns under CCN, it is facing the difficulty of choosing an optimal decision threshold for a scoring function that already is optimal.

Cohen et al. \cite{cohen2018machine} suggest to choose the decision threshold from known label noise rates. The authors discuss the implications of noise rate uncertainties from a physics perspective, concluding that small inaccuracies of the assumed noise rates do not harm the classifier training. This domain-specific result falls in line with the general CCN theory by Natarajan et al. \cite{natarajan2013learning}. If the noise rates of the full training set are unknown, they might at least be available for a small calibration set from which a decision threshold can be determined \cite{metodiev2017classification}.

The work by de Simone et al. \cite{desimone2019guiding} is mostly concerned with a two-sample hypothesis test. This test is used to decide whether there is \emph{some} signal in the data, with the classification of individual instances being only proposed as a future extension of their method. However, this extension would require a meaningful decision threshold to be set, just like in any other CCN task.

Two-sample hypothesis testing, in general, is concerned with the detection of differences between the distributions of two data samples \cite{gretton2012kernel}. Our learning criterion, the significance of detection, is motivated by one particular test which only checks whether the two \emph{On} and \emph{Off} samples differ in their numbers of positive instances. Other types of differences between the two data samples, which could be detected with other two-sample hypothesis tests, are not assumed here.

\section{Conclusion and Outlook}
\label{sec:conclusion}

We have proposed the significance of detection as a learning criterion for binary classification under class-conditional label noise. This criterion assumes one of the class-wise noise rates to be known, but does not need to estimate the other noise rate explicitly. We can use the criterion either to tune the decision threshold of any given classifier, or to learn a classification model from scratch. Most importantly, the significance of detection is interpretable in the sense that it indicates whether learning is actually feasible with the given noisy labels---a property that is established by an integrated hypothesis test. All of these qualities go beyond the state of the art in learning under class-conditional label noise. 

Our proposals stem from the gamma hadron classification task in gamma ray astronomy, where the significance of detection is already employed to tune the decision thresholds of strongly supervised models. These models are trained with simulated data, which is costly in terms of computation time and energy. Constrastingly, our weakly supervised algorithms learn directly from the real telescope, which produces an abundance of cheap but noisy-labeled data. Still, they match the performance of the strongly supervised state of the art in the gamma ray domain. We hope that this result in particular---and our discussion of gamma hadron classification as noisy label learning in general---will pave the way towards an even more data-driven gamma ray astronomy.

Future work should look into adapting the significance of detection for other data representations. A lot of engineering around strongly supervised classification went into the design of the numeric features \cite{Bockermann/etal/2015a} that we have continued to use here. Learning from the underlying sensor data of the telescope, however, would require some form of representation learning, e.g. using deep neural networks \cite{Buschjaeger/etal/2020a}. Looking into deep clustering approaches \cite{Caron/etal/2018 ,Asano/etal/2020a}, for instance, might yield performances similar to the ones presented here, but without the need for manual feature engineering.

 \section*{Acknowledgements}
This work has been supported by Deutsche Forschungsgemeinschaft (DFG) within the Collaborative Research Center SFB 876, "Providing Information by Resource-Constrained Data Analysis", project A1 and C3. \url{http://sfb876.tu-dortmund.de}. Parts of this work have been funded by the Federal Ministry of Education and Research of Germany as part of the competence center for machine learning ML2R (01|S18038A).

\bibliographystyle{IEEEtran}
\bibliography{main.bib}

\clearpage
\appendix

The following results provide further details regarding our experimental discussions from Sec.~\ref{sec:experiments}.

\begin{table}[!b]
\caption{Full version of Tab.~\ref{table:insample_results}, containing all hyper-parameter configurations that we have evaluated. We also compare to two other supervised models: a single decision tree that is trained with full supervision, and a random forest model ``Small RF'' that is trained on only 2,000 simulated examples.
}
\label{table:insample_results_full}
\centering
\addtolength{\tabcolsep}{-1pt}   
\begin{tabular}{@{}clcc@{}}
\toprule
\multicolumn{2}{l}{Model}           & Single Model & Ensemble $(T=100)$ \\ \midrule
\multirow{7}{*}{ \rotatebox[origin=c]{90}{\bf k-Means}} 
                                   & k=8               &$15.4230 \pm 0.3377$&$21.1536 \pm 0.3628$\\
                                   & k=16              &$18.5030 \pm 0.5808$&$22.1035 \pm 0.2753$\\
                                   & k=32              &$19.5769 \pm 0.6934$&$22.7916 \pm 0.2103$\\
                                   & k=48              &\textbf{19.5799 $\pm$ 0.3708}&$23.1176 \pm 0.1974$\\
                                   & k=64              &$18.7707 \pm 0.5221$&$23.3216 \pm 0.2562$\\
                                   & k=128             &$18.8068 \pm 0.4566$&$23.6613 \pm 0.4276$\\
                                   & k=256             &$17.7766 \pm 0.5713$&\textbf{24.3246 $\pm$ 0.4966}\\[3pt]
\multirow{5}{*}{ \rotatebox[origin=c]{90}{\bf LiMa-Tree}}                  
                                   & max\_depth=4      &\textbf{21.1666 $\pm$ 0.0000}&$23.3015 \pm 0.3321$\\
                                   & max\_depth=5      &$18.1350 \pm 0.0000$&$24.8517 \pm 0.2637$\\
                                   & max\_depth=6      &$18.4227 \pm 0.0000$&\textbf{25.1085 $\pm$ 0.2236}\\
                                   & max\_depth=7      &$16.4555 \pm 0.0000$&$24.9136 \pm 0.3942$\\
                                   & max\_depth=8      &$15.1137 \pm 0.0000$&$25.0431 \pm 0.2479$\\[3pt]                                    
\multirow{5}{*}{ \rotatebox[origin=c]{90}{\bf Noisy-Tree}}                  
                                   & max\_depth=4      &\textbf{24.2899 $\pm$ 0.0000}&$25.3015 \pm 0.3321$\\
                                   & max\_depth=5      &$23.5378 \pm 0.0000$&$25.8569 \pm 0.2667$\\
                                   & max\_depth=6      &$23.8007 \pm 0.0000$&$26.1597 \pm 0.2236$\\
                                   & max\_depth=7      &$22.5541 \pm 0.0000$&$26.2558 \pm 0.3942$\\
                                   & max\_depth=8      &$21.0170 \pm 0.0000$&\textbf{26.3175 $\pm$ 0.2479}\\[3pt] 
\multirow{5}{*}{ \rotatebox[origin=c]{90}{\bf Menon}} 
                                   & max\_depth=4      &\textbf{24.2899 $\pm$ 0.0000}&$25.2321 \pm 0.1812$\\
                                   & max\_depth=5      &$22.2190 \pm 0.0000$&$25.7456 \pm 0.2723$\\
                                   & max\_depth=6      &$21.9019 \pm 0.0000$&$25.9519 \pm 0.3090$\\
                                   & max\_depth=7      &$18.3020 \pm 0.0000$&\textbf{26.0165 $\pm$ 0.2950}\\
                                   & max\_depth=8      &$17.6470 \pm 0.0000$&$25.9726 \pm 0.3622$\\[3pt] 
\multirow{5}{*}{ \rotatebox[origin=c]{90}{\bf Mithal}}
                                   & max\_depth=4      &\textbf{24.2899 $\pm$ 0.0000}&$25.0858 \pm 0.2089$\\
                                  & max\_depth=5      &$23.4302 \pm 0.0000$&$25.7248 \pm 0.2606$\\
                                   & max\_depth=6      &$23.2471 \pm 0.0000$&$26.0736 \pm 0.2071$\\
                                   & max\_depth=7      &$20.1539 \pm 0.0000$&$26.3846 \pm 0.2246$\\
                                   & max\_depth=8      &$19.7144 \pm 0.0000$&\textbf{26.4465 $\pm$ 0.1804}\\[3pt]                                              
\multirow{2}{*}{ \rotatebox[origin=c]{90}{\bf Sup.}}                  
                                   & DT / RF      &$24.6506 \pm 0.0000$&$26.2526\pm 0.2024$\\
                                   & Small DT / RF    &$21.7351 \pm 0.0000$&$22.2584\pm 0.2951$\\ 
 \bottomrule 
\end{tabular}\end{table}

\subsection{Crab Nebula Detection under Violated Assumptions}
\label{sec:apx:violated}
The significance of detection intends to verify that the \emph{On} region contains significantly more truly-positive instances than any of the \emph{Off} regions. A necessary plausibility check is therefore to investigate the effects of violations of this assumption. To this end, we remove all instances from the \emph{On} region and declare one of the \emph{Off} regions as being a fake \emph{On} region. As expected, all our models return significances that are close to zero, see Table~\ref{table:fake_experiments}. Such small values help practitioners in detecting violations of their assumptions, like the one we have introduced artificially. Conversely, this property of the significance score can promote trust in the idea that high significance scores indeed indicate an excess of true positives in the \emph{On} region. In other words: the fact that we intend to maximize the significance of detection does not mean that we falsely over-estimate this criterion.

\begin{table}
\caption{Analysis on a synthetic dataset that violates the assumption of excess gamma particles in the on region. Our methods correctly identify that there is no significant source detection and outputs LiMa significances close to zero.}	
\label{table:fake_experiments}
\centering
\addtolength{\tabcolsep}{-1pt}   
\begin{tabular}{@{}lrr@{}}
\toprule
Model           & Single Model & Ensemble $(T=100)$ \\ \midrule
kMeans(k=32)                       &$0.0000 \pm 0.0000$&$0.2478 \pm 0.3544$\\
kMeans(k=128)                      &$0.0000 \pm 0.0000$&$0.3162 \pm 0.6247$\\[3pt]               
Tree(max\_depth=4)                 &$0.0000 \pm 0.0000$&$0.0435 \pm 0.0746$\\
Tree(max\_depth=6)                 &$0.0000 \pm 0.0000$&$0.0612 \pm 0.0425$\\
 \bottomrule 
\end{tabular}\end{table}

\subsection{Agreement with Supervised Models}
\label{sec:apx:agreement}
How different are noisy label models from models that are trained with full supervision? To answer this question, we compute the correlation between the gamma scores that are predicted by our noisy-label ensembles with the gamma scores that are predicted by the strongly supervised state-of-the-art. For single models ($T=1$), we compute correlations with crisp classifications because these models do not return real-valued gamma scores. In Tab.~\ref{table:agreement}, two different correlation measures, Pearson and Spearman, reveal that our models, particularly the ensembles, show a high correlation with the fully supervised models that are currently used for the FACT telescope. This result suggests that our weakly supervised models have indeed learned to distinguish gamma rays from hadronic particles, without ever accessing instances that are labeled as such.

\begin{table}[b!]
\caption{Agreement between our methods and the fully supervised state-of-the-art predictions, as measured via Pearson and Spearman correlations.}
\label{table:agreement}
\centering
\addtolength{\tabcolsep}{-1pt}   
\begin{tabular}{@{}lcc@{}}
\toprule
Model           & Pearson & Spearman \\ \midrule
kMeans(k=32, $T$=1)                         &$0.4265 \pm 0.0200$&$0.4048 \pm 0.0203$\\
kMeans(k=128, $T$=1)                        &$0.3255 \pm 0.0133$&$0.3153 \pm 0.0159$\\     
Tree(max\_depth=4, $T$=1)                   &$0.4965 \pm 0.0000$&$0.4678 \pm 0.0000$\\
Tree(max\_depth=6, $T$=1)                   &$0.4042 \pm 0.0000$&$0.3732 \pm 0.0000$\\[3pt] 
kMeans(k=32, $T$=100)                       &$0.7266 \pm 0.0071$&$0.7044 \pm 0.0074$\\
kMeans(k=128, $T$=100)                      &$0.6877 \pm 0.0119$&$0.6274 \pm 0.0188$\\      
Tree(max\_depth=4, $T$=100)                 &\textbf{0.7706 $\pm$ 0.0060}&\textbf{0.7364 $\pm$ 0.0106}\\
Tree(max\_depth=6, $T$=100)                 &$0.7401 \pm 0.0062$&$0.6806 \pm 0.0087$\\
 \bottomrule 
\end{tabular}\end{table}

\begin{figure}
  \includegraphics[width=\linewidth]{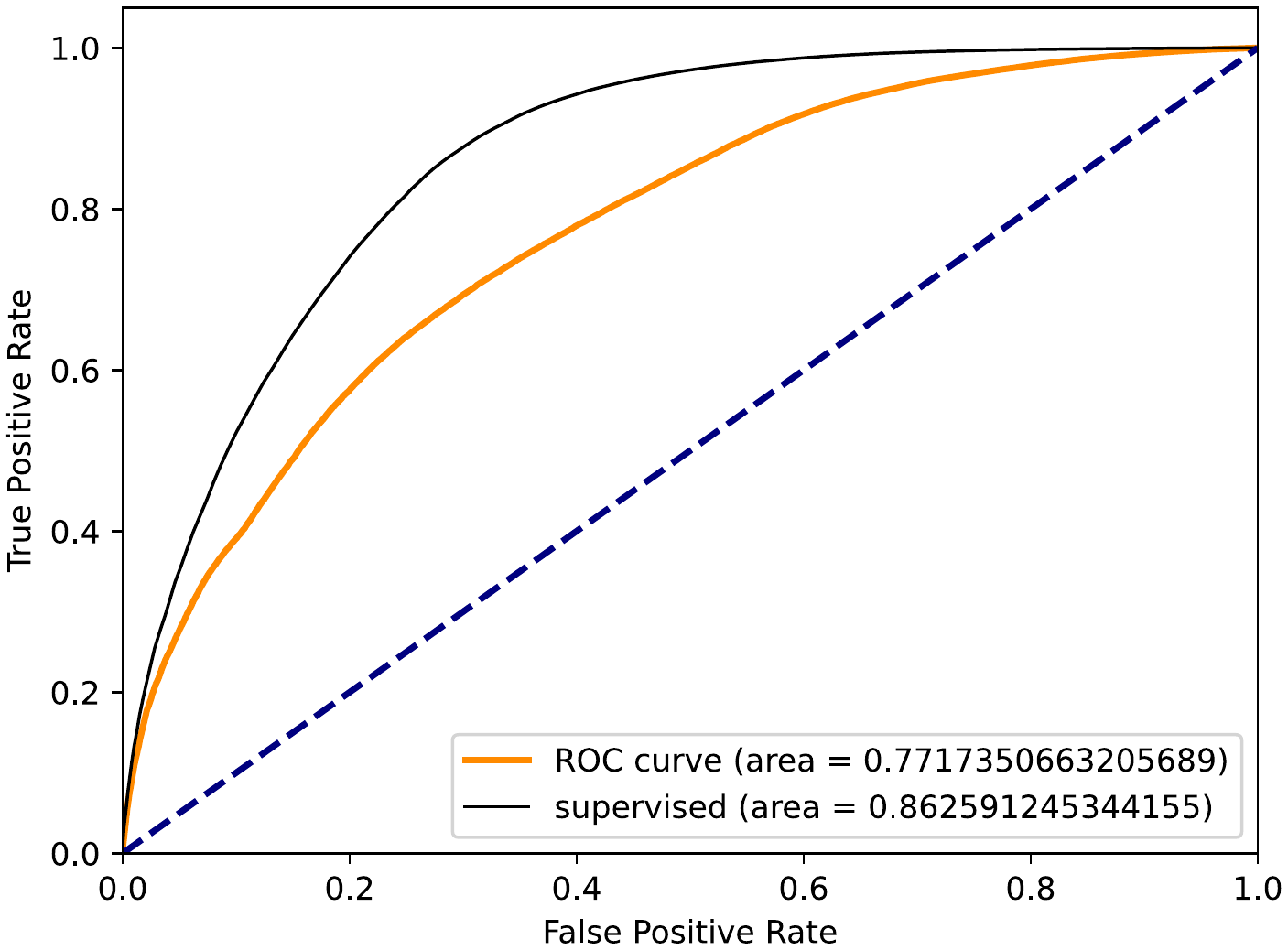}
  \caption{ROC curve comparing our LiMa Forest of depth 6 with the fully supervised state-of-the-art.}
  \label{fig:roc}
\end{figure}

As another matter of agreement, we test the performance of our noisy-label random forest on simulated ground-truth test data from which we can compute an ROC curve. As we can see in Figure~\ref{fig:roc}, our weakly supervised model achieves an area under the ROC curve of 0.77, which comes close to the fully supervised state-of-the-art. Note that the remaining difference between the two models is most likely due to deviations between the simulated and the real-world data. In fact, the simulation has its own imperfections. From a CCN learning stand-point, see Sec.~\ref{sec:background}, we know that models trained with class-conditional label noise have the same ROC curve as models trained with clean labels, at least asymptotically. Therefore, we can expect an optimal ROC when learning from a large dataset with class-conditional label noise; our clear CCN interpretation of \emph{On} and \emph{Off} information supports such a high expectation. In contrast to the \emph{On} and \emph{Off} noisy label interpretation, we are not aware of a similarly clear picture of the noise that is induced by an imperfect simulation.

\subsection{Results in Other Application Domains (Continued)}
\label{sec:apx:other}
We have already seen competitive performances of our weakly supervised models in imbalanced application domains. Namely, Tab.~\ref{table:classification_results} and Fig.~\ref{fig:ccd} already display results that are obtained by artificially introducing label noise with the rates $p_+ = 0.1$ and $p_- = 0.5$.

We have repeated this experiment with stronger label noise, namely with $p_+ = 0.25$. The results of this variation are displayed in Tab.~\ref{table:classification_results_more_noise}. As we can see, these results are in line with the results that are shown earlier for $p_+ = 0.1$. However, the overall magnitude of F1 scores is lower here.

\begin{table}
\caption{Averaged F1 scores on imbalanced data sets with noise levels $p_+=0.25$ and $p_-=0.5$.}
\label{table:classification_results_more_noise}
\centering
\addtolength{\tabcolsep}{-3pt}   
\scriptsize
\begin{tabular}{lcccc}
\toprule
Dataset        & LiMa Forest         & Noisy Forest        & Menon Forest         & Mithal Forest       \\ \midrule
satimage       & 0.445 $\pm$ 0.028   & \textbf{0.462 $\pm$ 0.023} & 0.456 $\pm$ 0.004  & 0.246 $\pm$ 0.029 \\
proteinhomo    & 0.767 $\pm$ 0.008   & 0.773 $\pm$ 0.004 & 0.198 $\pm$ 0.006  & \textbf{0.777 $\pm$ 0.004} \\
opticaldigits  & 0.556 $\pm$ 0.131   & \textbf{0.701 $\pm$ 0.016} & 0.504 $\pm$ 0.008  & 0.662 $\pm$ 0.081 \\
pendigits      & 0.872 $\pm$ 0.027   & 0.832 $\pm$ 0.010 & 0.900 $\pm$ 0.004  & \textbf{0.918 $\pm$ 0.004} \\
letterimg      & 0.710 $\pm$ 0.019   & \textbf{0.779 $\pm$ 0.007} & 0.539 $\pm$ 0.010  & 0.476 $\pm$ 0.233 \\
coil2000       & 0.104 $\pm$ 0.010   & 0.128 $\pm$ 0.008 & \textbf{0.161 $\pm$ 0.002}  & 0.113 $\pm$ 0.001 \\
thyroidsick    & \textbf{0.666 $\pm$ 0.016}   & 0.628 $\pm$ 0.020 & 0.509 $\pm$ 0.010 & 0.159 $\pm$ 0.025  \\
sickeuthyroid  & \textbf{0.661 $\pm$ 0.019}   & 0.589 $\pm$ 0.030 & 0.476 $\pm$ 0.009 & 0.256 $\pm$ 0.032  \\
mamography     & \textbf{0.492 $\pm$ 0.020}   & 0.418 $\pm$ 0.021 & 0.238 $\pm$ 0.005 & 0.051 $\pm$ 0.005  \\
abalone        & \textbf{0.327 $\pm$ 0.011}   & 0.312 $\pm$ 0.016 & 0.281 $\pm$ 0.004 & 0.173 $\pm$ 0.004  \\[1pt]
mean rank      & 2.200               & \textbf{1.900}       & 2.800               & 3.100           \\
mean f1        & \textbf{0.566}      & 0.562                & 0.426               & 0.3829           \\
\bottomrule
\end{tabular}

\end{table}

\end{document}